\documentclass[letterpaper, 10 pt, journal, twoside]{IEEEtran}
\ifCLASSINFOpdf
\else
\fi
\usepackage{multicol}
\usepackage[bookmarks=true]{hyperref}
\usepackage{stix}
\usepackage{amsmath} 
\usepackage{amssymb}  
\usepackage{xcolor}
\usepackage{xspace}
\usepackage{filecontents}

\definecolor{ceil}{rgb}{0.16, 0.32, 0.75}
\definecolor{ceruleanblue}{rgb}{0.16, 0.32, 0.75}
\definecolor{chestnut}{rgb}{0.8, 0.36, 0.36}

\newcommand{\sz}[1]{\textcolor{red}{\bf\small }}
\newcommand{\sarvesh}[1]{\textcolor{green}{\bf\small }}

\newcommand{\apdx}[1]{\href{https://xianyicheng.github.io/HiDex-Website/static/files/HiDex-Appendix.pdf}{our appendices}}

\usepackage[normalem]{ulem}

\usepackage{verbatim}

\usepackage{algpseudocode}
\usepackage{graphicx}
\usepackage{adjustbox}
\usepackage{tikz}
\usepackage{pgfplots}
\usepackage{textcomp}
\usepackage{xcolor}
\usepackage{booktabs}
\usepackage{siunitx}

\usepackage{multirow,array}

\usepackage{subcaption}
\usepackage{hyperref}
\usepackage{footmisc}
\usepackage{subfloat}
\usepackage[utf8]{inputenc}
\usepackage[english]{babel}
\usepackage{epstopdf}
\usepackage{amsthm}
\theoremstyle{definition} 
\newtheoremstyle{italdefinition} 
{1pt} 
{1pt} 
{} 
{10pt} 
{\itshape} 
{:} 
{3pt} 
{} 
\theoremstyle{italdefinition}
\newtheorem{definition}{Definition}

\usepackage[normalem]{ulem} 
\usepackage{xcolor}

\newcommand{\redtext}[1]{\textcolor{black}{#1}}
\newcommand{\bluetext}[1]{\textcolor{black}{#1}}
\newcommand{\purpletext}[1]{\textcolor{black}{#1}}
\newcommand{\greentext}[1]{\textcolor{black}{#1}}

\newcommand{\parallelsum}{\mathbin{\!/\mkern-5mu/\!}}

\usepackage[linesnumbered,ruled,vlined]{algorithm2e}
\usepackage{setspace}

\usepackage{ifthen}
\newboolean{includeFigures}
\setboolean{includeFigures}{true}

\usepackage[font=small,skip=3pt]{caption}

\let\oldtwocolumn\twocolumn
\renewcommand\twocolumn[1][]{%
    \oldtwocolumn[{#1}{
    \begin{center}
           \includegraphics[width=\textwidth]{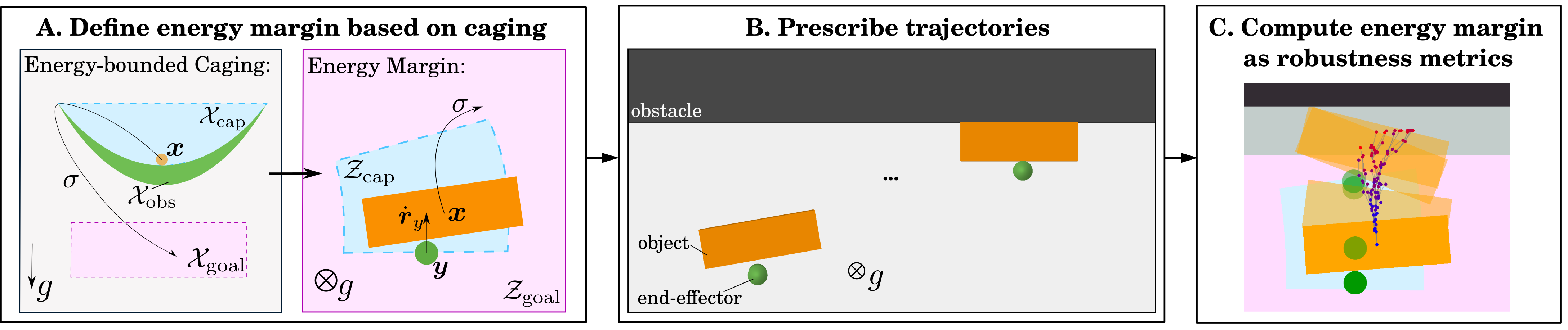}
           \captionof{figure}{\purpletext{Inspired by caging analysis (A, left), we propose to characterize the robustness of manipulation actions using energy margins (A, right) from failure regions (pink). Given prescribed trajectories of manipulation tasks (B, planar pushing as an example), we use sampling-based kinodynamic planners to compute the metrics (C), which are used to predict manipulation robustness and success under uncertainty.}}
           \label{fig-sum}
        \end{center}
    }]
}

\usepackage{titlesec}
\titlespacing*{\section}{3pt}{6pt}{2pt}
\titlespacing*{\subsection}{3pt}{6pt}{2pt}
\usepackage{indentfirst}

\begin{document}
%
\title{Characterizing Manipulation Robustness through Energy Margin and Caging Analysis}
%
%
%

\author{Yifei Dong$^1$, Xianyi Cheng$^2$ and Florian T. Pokorny$^1$%
\thanks{Manuscript received: March 26, 2024; Accepted: May 21, 2024.} 
\thanks{This paper was recommended for publication by Editor-in-Chief Tamim Asfour upon evaluation of the Editor, Associate Editor and Reviewers’ comments.
This work was partially funded by the European Commission under the Horizon Europe Framework Programme project \href{https://softenable.eu/}{SoftEnable}, grant number 101070600.} 
\thanks{$^1$The authors are with the division of Robotics, Perception and Learning, KTH Royal Institute of Technology, 10044 Stockholm, Sweden, {\tt\small \{yifeid, fpokorny\}@kth.se}.}
\thanks{$^2$The author is with Carnegie Mellon University, Pittsburgh, PA, 15213, USA. {\tt\small xianyic@andrew.cmu.edu}.}
\thanks{Digital Object Identifier (DOI): see top of this page.}
}

%
%

\markboth{IEEE Robotics and Automation Letters. Preprint Version. June, 2024}
{Dong \MakeLowercase{\textit{et al.}}: Characterizing Manipulation Robustness through Energy Margin and Caging Analysis} 

%



\maketitle

\begin{abstract}
To develop robust manipulation policies, quantifying robustness is essential. Evaluating robustness in general manipulation, nonetheless, poses significant challenges due to complex hybrid dynamics, combinatorial explosion of possible contact interactions, global geometry, etc. This paper introduces an approach for evaluating manipulation robustness through energy margins and caging-based analysis. Our method assesses manipulation robustness by measuring the energy margin to failure and extends traditional caging concepts for dynamic manipulation. This global analysis is facilitated by a kinodynamic planning framework that naturally integrates global geometry, contact changes, and robot compliance. We validate the effectiveness of our approach in simulation and real-world experiments of multiple dynamic manipulation scenarios, highlighting its potential to predict manipulation success and robustness.
\end{abstract}











%
\IEEEpeerreviewmaketitle

\section{Introduction}
\label{sec-intro}
Human manipulation is intriguing due to its dexterity, simplicity, and remarkable robustness. Consider picking up a thin object from a table: this involves complex interactions between the object, table, and hand, yet it remains robust and fast. The robustness of human manipulation is often attributed not to precise control over or knowledge of contact interactions but to factors like hand kinematic configurations, environment geometry, joint compliance, material softness, and so on. An empirical study in \cite{bhatt2022surprisingly} discovers similar robustness in in-hand manipulation with a soft robotic hand. 

Nonetheless, there is a lack of methods in the literature for evaluating such general robustness. Existing research studying similar robustness problems can generally be categorized into three areas: grasp quality metrics, control stability analysis (including contraction analysis \cite{tsukamoto2021contraction}, Lyapunov functions \cite{tedrake2009lqr} and barrier functions \cite{ames2019control}), and caging. Most quality metrics in grasping \cite{roa2015grasp} and non-prehensile manipulation \cite{hou2019criteria} often define robustness through object position or wrench space margins based on first-order analysis. Control stability methods assess robustness by examining system evolution from initial states, offering a local perspective that struggles to account for the robustness under the combinatorial explosion of possible contact interactions, the resulting hybrid dynamics, and global geometric properties. In contrast, caging approaches, which confine an object within a bounded space to prevent escape, offer a more global geometric analysis of robustness.

This paper proposes an approach to characterize robustness in manipulation via energy margins (Fig. \ref{fig-sum}). Our approach expands on prior quasi-static cage \cite{makita2017survey} and soft fixture \cite{dong2023quasi} analyses by adapting classical caging concepts to a kinodynamic context, where dynamically feasible escape paths are considered. Utilizing forward simulation enables natural consideration of contact changes, robot joint stiffness, complex geometries, and system kinodynamics. This methodology introduces a new lens by evaluating the robustness of manipulation strategies through energy margins and broadens the applicability of caging concepts to dynamic scenarios. 

The main contributions of this paper are: 
1. We propose to characterize more general robustness in manipulation with energy margins and caging analysis. This approach evaluates manipulation strategies that consider global geometry, kinodynamics, and complex contact events.
2. We develop a sampling-based kinodynamic planning framework for dynamic caging analysis and energy margin calculation, designed to be inherently efficient in dynamic, contact-rich environments.
3. Through experimental validation in both real-world and simulated settings, we demonstrate the practicality and effectiveness of our approach, showcasing its benefits over a wrench-based baseline method.

\section{Related Work}
\subsection{Caging}
Classical caging focuses on preventing an object from escaping using geometric constraints. This foundational concept, as introduced by Kuperberg \cite{kuperberg1990problems} and further developed by Rimon et al. \cite{rimon1996caging}, \cite{rimon1999caging}, \bluetext{Maeda et al. \cite{maeda2017sensorless}}, explores how robots can effectively constrain objects, without necessitating force closure or form closure. Rodriguez et al. \cite{rodriguez2012caging} demonstrate utilizing caging to achieve immobilizing grasps. Energy-bounded caging \cite{mahler2016energy, mahler2018synthesis} relaxes complete geometric caging and considers the work done by external forces, such as gravity \cite{mahler2016energy} and planar friction \cite{mahler2018synthesis}, as a source to cage objects. 
\greentext{
Caging has also been leveraged in the robust manipulation of biological cells using optical tweezers \cite{thakur2014indirect,ta2020caging}.}
Our previous work on soft fixtures \cite{dong2023quasi} presents sampling-based methods to search for upper-bound estimates of escape energy in higher-dimensional configuration spaces, such as simplified deformable objects.
Our approach presents a new perspective that enables caging analysis to be used to characterize general manipulation robustness. It also extends caging into dynamic settings, bridging the gap between static caging theories and the dynamic nature of robotic manipulation. 

\subsection{Robustness and Quality Metrics}
Most research in manipulation robustness focuses on grasp quality measures.
A comprehensive review by Roa et al. \cite{roa2015grasp} categorizes classical quality measures based on contact configurations and wrench space profiles. 
Many metrics are derived from Grasp Wrench Space (GWS) \cite{pollard1996synthesizing}, which constructs a wrench cone formed by grasping contacts, such as the volume of the GWS \cite{miller1999examples}, the largest perturbation wrench the grasp can resist in any direction \cite{ferrari1992planning}, etc. 
Lin et al. \cite{lin2015task} introduce a task-specific grasp quality criterion based on the distribution of task disturbance, exemplifying the importance of task-specific optimization of grasps. Xu et al. \cite{xu2020minimal} propose minimal work as a grasp quality metric to plan grasps for 3D deformable hollow objects considering wrench resistance. Machine learning approaches are also employed to predict grasp success based on data labels derived from analytic quality metrics \cite{saxena2008learning}. 
For non-prehensile manipulation, a framework of quasi-static wrench-based analysis is presented for balancing grasps in \cite{krug2018evaluating}. Makapunyo et al. \cite{makapunyo2013measurement} employ sampling-based methods to determine a quality metric for partial caging. 
While previous work mostly focuses on first-order and instantaneous analysis to quantify how good an action is under position or object wrench uncertainty, we use caging to characterize robustness through a more global analysis, with the ability to take into account many hard-to-model effects, such as contact changes and friction uncertainty. 
\section{Preliminaries}
\subsection{Nomenclature}
We denote by $\mathcal{X} \subset \mathrm{SE}(3)$ the configuration space (C-space) of a 3D rigid object. The set of time derivatives of the object configuration is denoted by $\dot{\mathcal{X}} \subset T\mathrm{SE}(3)$, where $T\mathrm{SE}(3)$ is the tangent space associated with the Lie group $\mathrm{SE}(3)$. 
The C-space of a robot end-effector and surrounding static obstacles is denoted by $\mathcal{Y} \subset \mathrm{SE}(3) \times \mathbb{R}^{n_r}$, where $\mathrm{SE}(3)$ denotes the base pose, and $n_r$ refers to the number of joints. Its set of time derivatives is denoted by $\dot{\mathcal{Y}} \subset T\mathrm{SE}(3) \times \mathbb{R}^{n_r}$. $\mathcal{Y}$ might degenerate when, for example, the base of the end-effector is fixed, and then $\mathcal{Y} \subset \mathbb{R}^{n_r}$. For planar manipulation, $\mathcal{X}$ and $\mathcal{Y}$ degenerate to $\mathrm{SE}(2)$ and $\mathrm{SE}(2) \times \mathbb{R}^{n_r}$.
An element $\boldsymbol{x}\in\mathcal{X}$, described as 
$\boldsymbol{x}=(\boldsymbol{r}_x, \boldsymbol{q}_x)$, encompasses the position $\boldsymbol{r}_x$ of the Center of Mass (CoM) and orientation $\boldsymbol{q}_x$ (a unit quaternion). An element $\dot{\boldsymbol{x}} \in \dot{\mathcal{X}}$ is described as 
$\dot{\boldsymbol{x}} =(\dot{\boldsymbol{r}}_x, \dot{\boldsymbol{q}}_x)$, 
which comprises the CoM linear velocity $\dot{\boldsymbol{r}}_x$ and the derivative of quaternion $\dot{\boldsymbol{q}}_x$. Similarly for ${\boldsymbol{y}} \in \mathcal{Y}$ and $\dot{\boldsymbol{y}} \in \dot{\mathcal{Y}}$, we have $\boldsymbol{y}=(\boldsymbol{r}_y, \boldsymbol{q}_y, \boldsymbol{\alpha}_y)$ and $\dot{\boldsymbol{y}} =(\dot{\boldsymbol{r}}_y, \dot{\boldsymbol{q}}_y, \dot{\boldsymbol{\alpha}}_y)$, where $\boldsymbol{\alpha}_y$ and $\dot{\boldsymbol{\alpha}}_y$ refer to the joint position and velocity of the end-effector. \bluetext{$\boldsymbol{r}_y$ and $\boldsymbol{q}_y$ denote the position and orientation of the end-effector.}

\subsection{Cage and Soft Fixture}
Here, we revisit the concept of caging. The free C-space $\mathcal{X}_{\text{free}}$ indicates a set of configurations for which the object does not penetrate any of the bodies (the end-effector or obstacles) in the workspace.

\begin{definition}
    A \textit{cage} occurs when an object configuration $\boldsymbol{x}_{\text{init}}\in \mathcal{X}_{\text{free}}$ is situated in a bounded path component of $\mathcal{X}_{\text{free}}$.
    \label{def1}
\end{definition}

An object in a cage thus indicates its limited capacity to freely move beyond a certain proximity to its starting configuration. Caging was relaxed as partial caging by allowing escape paths through narrow passages in the free space \cite{makapunyo2013measurement}. Partial caging further extends to energy-bounded caging \cite{mahler2016energy} and soft fixtures \cite{dong2023quasi} that consider both geometric and potential energy constraints, as introduced below.

\begin{definition}
    In scenarios where only conservative forces act within a quasi-static system, an object at an initial configuration $\boldsymbol{x}_{\text{init}}\in\mathcal{X}_{\text{free}}$ possesses potential energy $E(\boldsymbol{x}_{\text{init}})$. Its \emph{escape energy} $\mathcal{E}(\boldsymbol{x}_{\text{init}})$ refers to the supremum of $e$ such that the path component $\mathcal{PC}_{\boldsymbol{x}_{\text{init}}} (\mathcal{X}_{e}(\boldsymbol{x}_{\text{init}}))$ of a sublevel set $\mathcal{X}_{e}(\boldsymbol{x}_{\text{init}})$ containing $\boldsymbol{x}_{\text{init}}$ is bounded:
    \begin{align} 
    \mathcal{E}(\boldsymbol{x}_{\text{init}}) &= \sup\left\{e \ge 0 : \mathcal{PC}_{\boldsymbol{x}_{\text{init}}} (\mathcal{X}_{e}(\boldsymbol{x}_{\text{init}}))  \text{ is bounded} \right\}, \label{eq0}
    \end{align}
    where
    \begin{equation} \label{eq1}
    \mathcal{X}_{e}(\boldsymbol{x}_{\text{init}}) = \{\boldsymbol{x} \in \mathcal{X}_{\text{free}} : E(\boldsymbol{x}) \le E(\boldsymbol{x}_{\text{init}}) + e \}.
    \end{equation}
    The object is deemed to be in a \emph{soft fixture} if $\mathcal{E}(\boldsymbol{x}_{\text{init}}) > 0$. Note that $\mathcal{E}(\boldsymbol{x}_{\text{init}})$ is defined only if
    $\mathcal{PC}_{\boldsymbol{x}_{\text{init}}} (\mathcal{X}_{0}(\boldsymbol{x}_{\text{init}}))$ is bounded.
    \label{def2}
\end{definition}

In practice, sampling-based approaches can approximate a close upper bound of $\mathcal{E}(\boldsymbol{x}_{\text{init}})$ by utilizing a goal region (Fig. \ref{fig-sum}-A, left) sufficiently far away from the obstacles with relatively lower energy values and searching for geometrically feasible escape paths with lowest energy cost \cite{dong2023quasi}.

\subsection{Assumptions}
\label{subsec-assume}
We introduce the assumptions that this paper relaxes compared to our prior work \cite{dong2023quasi}.

\textit{1) System Motion}: 
While the previous work only considers quasi-static conditions, here we account for the movement of both the object and the end-effector, while keeping other environmental elements static. We introduce $\mathcal{Z} = \mathcal{X} \times \dot{\mathcal{X}} \times \mathcal{Y} \times \dot{\mathcal{Y}}$, where $\mathcal{Z}_{\text{free}}$ represents penetration-free space. In this context, the energy function $E$ is the system's total mechanical energy (sum of kinetic and potential energy).

\textit{2) Non-Conservative Forces}: 
Given the existence of non-conservative forces, such as friction, the previous definition of escape energy in Def. \ref{def1} is problematic. Consider the example of planar pushing (Fig. \ref{fig-sum}-A, right), assuming the tabletop with a rough surface is infinitely large. The escape energy $\mathcal{E}(\boldsymbol{x}_{\text{init}})$ is always infinite in this scenario because the object overcomes friction along any escape path to a point that is arbitrarily far away. To utilize escape energy for more general cases like this, we define a \textit{capture set} $\mathcal{Z}_{\text{cap}}(\boldsymbol{z}_{\text{init}}) \subset \mathcal{Z}$ instead, 
\begin{equation}
    \redtext{ \mathcal{Z}_{\text{cap}}(\boldsymbol{z}_{\text{init}}) = \{\boldsymbol{z} \in \mathcal{Z} : f(\boldsymbol{z}) \leq c_1(\boldsymbol{z}_{\text{init}}), g(\boldsymbol{z}) = c_2(\boldsymbol{z}_{\text{init}})\}. }
\end{equation}
\redtext{
The constraints above are task-specific, refraining the system from transitioning to failure modes. 
For instance, in the pushing task (Fig. \ref{fig-sum}-A, right), $\mathcal{Z}_{\text{cap}}(\boldsymbol{z}_{\text{init}})$ contains positional constraints that prevent the object from slipping away from the end-effector.
A system state $\boldsymbol{z} \in \mathcal{Z}_{\text{cap}}(\boldsymbol{z}_{\text{init}})$ indicates the object as dynamically controllable by the end-effector for achieving specific tasks. The effort of the object to escape from $\mathcal{Z}_{\text{cap}}(\boldsymbol{z}_{\text{init}})$ thus indicates manipulation robustness, implying an energy margin from failure. Several examples of task-specific $\mathcal{Z}_{\text{cap}}(\boldsymbol{z}_{\text{init}})$ can be found in Section \ref{subsec-examples} and Fig. \ref{fig-examples}.
}

\textit{3) Dynamic Escape Paths}: 
We consider dynamically feasible escape paths that lead the object out of the capture set $\mathcal{Z}_{\text{cap}}(\boldsymbol{z}_{\text{init}})$. 
A dynamically feasible path $\sigma: [0, 1] \to \mathcal{Z}_{\text{free}}$, such that $\sigma(0)=\boldsymbol{z}_{\text{init}} $ and $ \sigma(1)=\boldsymbol{z}_{\text{goal}}$, is generated by applying random wrenches to the object to mimic perturbation or uncertainty during the manipulation. Here, $\boldsymbol{z}_{\text{goal}} \in \mathcal{Z}_{\text{goal}}$ and $\mathcal{Z}_{\text{goal}}$ is a goal set of escape. 

Besides the assumptions above, we only consider rigid objects under potential energy fields, though our framework can potentially be extended to scenarios with articulated or multiple objects with more computational costs caused by the increased degrees of freedom of the system.

\subsection{Problem Statement}
We tackle the following problem: Given a manipulation task objective and the state $\boldsymbol{z}_{\text{init}}$ of a rigid object and an end-effector in motion, which is subject to the assumptions in Section \ref{subsec-assume}, 1. Define metrics to characterize robustness in manipulation based on energy margins to failure (Section \ref{sec-def}). 2. Compute the robustness metrics using kinodynamic motion planning and evaluate their performance (Section \ref{sec-algo} and \ref{sec-eval}).

\section{Energy Margin Definitions}
\label{sec-def}

We quantify robustness with two methods based on energy margin: the effort of escape (the ease of failure) and the capture score (the probability of staying safe).

\subsection{Effort of Escape}
\label{subsec-escape}
The effort of escape calculates the minimal effort required to exit the capture set.
\begin{definition}
    The minimal \emph{effort of escape} is the minimal integral of the absolute power $\lvert\dot{W}_{\text{ext}}(\sigma(t))\rvert$ of the external work $W_{\text{ext}}(\sigma(t))$ required to be done to help the object escape from the capture set $\mathcal{Z}_{\text{cap}}(\boldsymbol{z}_{\text{init}})$:
    \begin{equation} \label{eq2}
    \Omega_{\text{esc}}\left(\boldsymbol{z}_{\text{init}}\right) = \min_{\sigma \in \Sigma(\boldsymbol{z}_{\text{init}}, \mathcal{Z}_{\text{goal}})} \int_{0}^{1} \lvert \dot{W}_{\text{ext}}(\sigma(t)) \rvert dt,
    \end{equation}
    where
    \begin{equation} \label{eq3}
    \dot{W}_{\text{ext}}(\sigma(t)) = \frac{dE(\sigma(t))}{dt} - \dot{W}_{\text{fri}}(\sigma(t)).
    \end{equation}
    \label{def3}
    $\frac{dE(\sigma(t))}{dt}$ is the instantaneous rate of change of the mechanical energy of the system, and $W_{\text{fri}}(\sigma(t))$  represents the work done by friction at the instantaneous state of the system. We denote $\Sigma(\boldsymbol{z}_{\text{init}}, \mathcal{Z}_{\text{goal}})$ as the set of all escape paths starting from $\boldsymbol{z}_{\text{init}}$ terminating in $\mathcal{Z}_{\text{goal}}$, where $\mathcal{Z}_{\text{goal}} = \mathcal{Z}_{\text{cap}}^c(\boldsymbol{z}_{\text{init}})$ (the complement set of the capture set).
\end{definition}
Eq. \eqref{eq3} follows \purpletext{the law of conservation of energy}.
We employ the absolute value of the power $\lvert \dot{W}_{\text{ext}}(\sigma(t)) \rvert$ as a measure of the total control input that we want to minimize because, for example, both speeding up and slowing down a vehicle require fuel, and we might want to minimize the total fuel used over a journey.
Since paths in the path space $\Sigma(\boldsymbol{z}_{\text{init}}, \mathcal{Z}_{\text{goal}})$ can hardly be enumerated or optimized analytically given the complexity introduced by friction, we employ kinodynamic sampling-based planners to filter candidate paths that upper bounds $\Omega_\text{esc}\left(\boldsymbol{z}_{\text{init}}\right)$.

\subsection{Capture Score}
\label{subsec-likelihood}
The capture score estimates the likelihood of remaining within the capture set. 
We consider an energy cost field and its correlating probabilistic distribution of the system. Specifically, a sequence of random wrenches applied on the object at $\boldsymbol{z}_{\text{init}}$ lead it to $\boldsymbol{z}^{0}$ with a cost of total effort $c^0$. By repeating the process $M$ times from $\boldsymbol{z}_{\text{init}}$, the system terminates in a list of states $\{\boldsymbol{z}^{0}, ..., \boldsymbol{z}^{M}\}$ with corresponding costs $\{c^0,...,c^M\}$. Data pairs in $\{(\boldsymbol{z}^{0}, c^0), ..., (\boldsymbol{z}^{M}, c^M)\}$ constitute an energy cost field that demonstrates the state space reachability in terms of energy cost from $\boldsymbol{z}_{\text{init}}$.
\redtext{We provide a probabilistic interpretation of the likelihood of reaching each state $\boldsymbol{z}^{m} \in \mathcal{Z}$ by introducing a probability mass function $L: \mathcal{Z} \to \mathbb{R}^{+}$,}
\begin{equation}\label{eq4}
    L(\boldsymbol{z}^{m}) = \frac{e^{-\lambda(c^{m}-c_{\text{min}})}}{\sum_{m=0}^{M} e^{-\lambda(c^{m}-c_{\text{min}})}},
\end{equation}
where $c_{\text{min}} = \min_{0 \leq m \leq M}{c^{m}}$ and $\lambda$ is a hyper-parameter. \redtext{The softmax function in Eq. (\ref{eq4}) is ideal for modeling hybrid continuous-to-discrete mappings in such stochastic systems \cite{ahmed2012bayesian}.}
\redtext{The approximated likelihood function indicates some states with lower costs are probabilistically more reachable.}

\begin{definition}
    The \textit{capture score} is defined as the sum of likelihood values of samples in a capture set $\mathcal{Z}_{\text{cap}}(\boldsymbol{z}_{\text{init}})$,
    \begin{equation} \label{eq5}
        \Omega_{\text{cap}}(\boldsymbol{z}_{\text{init}}) = \sum_{m=0}^{M} \delta(\boldsymbol{z}^{m} \in \mathcal{Z}_{\text{cap}}(\boldsymbol{z}_{\text{init}})) \cdot L(\boldsymbol{z}^{m}),
    \end{equation}
    where $\delta(\cdot)$ is an indicator function that equals $1$ if the condition inside the brackets is satisfied and $0$ otherwise. 
    \label{def4}
\end{definition}
Similarly, we define a task success set $\mathcal{Z}_{\text{suc}} \subset \mathcal{Z}$ independent of $\boldsymbol{z}_{\text{init}}$, indicating the set of states symbolizing the object accomplishes the manipulation objective. A capture score of success is thus given by
\begin{equation} \label{eq6}
    \Omega_{\text{suc}}(\boldsymbol{z}_{\text{init}}) = \sum_{m=0}^{M} \delta(\boldsymbol{z}^{m} \in \mathcal{Z}_{\text{suc}}) \cdot L(\boldsymbol{z}^{m}),
\end{equation}
which is essentially a predictor for fulfilling a task-specific objective from the state $\boldsymbol{z}_{\text{init}}$. In practice, we employ kinodynamic motion planners rather than repetitive Monte Carlo rollouts from $\boldsymbol{z}_{\text{init}}$. Sampling-based kinodynamic motion planners are in general more time- and memory-efficient by using strategies such as biased sampling and caching explored nodes.

\section{Computing Energy Margin Through Kinodynamic Planning}
\label{sec-algo}

We develop kinodynamic motion planning algorithms to compute the scores in Section \ref{sec-def}. We calculate the effort of escape $\Omega_{\text{esc}}(\boldsymbol{z}_{\text{init}})$ by an iterative tree search algorithm (Section \ref{V1}). We calculate the capture scores $\Omega_{\text{cap}}(\boldsymbol{z}_{\text{init}})$ and $\Omega_{\text{suc}}(\boldsymbol{z}_{\text{init}})$ by growing an expansive tree and approximating its correlating probabilistic distribution (Section \ref{V2}).  

\subsection{Planning Objectives} \label{V0}
We consider a system with an object, an end-effector, and the environment. 
The system state at time $k$ is 
\begin{equation} \label{eq7}
\boldsymbol{z}(k) = \left[\begin{array}{llll}\boldsymbol{x}^{\top}(k) & \dot{\boldsymbol{x}}^{\top}(k) & \boldsymbol{y}^{\top}(k) & \dot{\boldsymbol{y}}^{\top}(k) \end{array}\right]^{\top}. 
\end{equation} 
A manipulation trajectory is denoted by $\{(k,\boldsymbol{z}(k)): k \in [0,K]\}$, complying with the system dynamics ${\boldsymbol{z}}^{\prime}(k) = F\left(\boldsymbol{z}(k), \boldsymbol{\phi}(k)\right)$. $\boldsymbol{\phi}$ denotes the robot control. The system dynamics is rolled out through physics simulation. For each state $\boldsymbol{z}(k)$ along the trajectory, Algo. \ref{algo-escape} computes the minimal effort of escape $\Omega_{\text{esc}}(\boldsymbol{z}(k))$, and Algo. \ref{algo-likelihood} computes the capture scores $\Omega_{\text{cap}}(\boldsymbol{z}(k))$ and $\Omega_{\text{suc}}(\boldsymbol{z}(k))$. \redtext{We assume that the end-effector is not actuated in the short period when computing the metrics so as not to introduce extra energy sources to the system.} 

\begin{algorithm}[tp]
\small
\text{Initialize}, $P$, $A$, $n$ \\
$\sigma_0 \gets A(P_{\infty})$ \hfill {\footnotesize $\triangleright$ \textcolor{blue}{\text{Find a first escape path}}} \\
\If{$\sigma_0$ \text{does not exists}} 
    {$P$ \text{has no solution}\\
    \textbf{Return} $\infty$ \hfill {\footnotesize $\triangleright$ \textcolor{blue}{\text{Infinite {effort} to escape}}} \\
    } 
$c_0 \gets C(\sigma_0)$ \hfill {\footnotesize $\triangleright$ \textcolor{blue}{\textsc{eq. \eqref{eq8}.}} \textcolor{blue}{\text{Initial {effort of escape}}}}\\
\For{$i=1,2, \ldots, n$}{
    $\sigma_i \gets A(P_{c_{i-1}})$ \hfill {\footnotesize $\triangleright$ \textcolor{blue}{\text{Find a path of lower {effort}}}} \\
    $c_i \gets C(\sigma_i)$ \hfill {\footnotesize $\triangleright$ \textcolor{blue}{\textsc{eq. \eqref{eq8}.}} \textcolor{blue}{\text{Update {effort} upper bound}}} \\
    }
$\Omega_{\text{esc}}(\boldsymbol{z}_{\text{init}}) \gets c_n$ \hfill {\footnotesize $\triangleright$ \textcolor{blue}{\textsc{eq. \eqref{eq2}.}} \textcolor{blue}{\text{Final {effort of escape}}}} \\
\textbf{Return} $\Omega_{\text{esc}}(\boldsymbol{z}_{\text{init}})$
\caption{Compute Effort of Escape}
\label{algo-escape}
\end{algorithm}

\subsection{Computing Effort of Escape} \label{V1}
We frame finding the minimum effort to escape from the capture set $\mathcal{Z}_{\text{cap}}(\boldsymbol{z}_{\text{init}})$ as an optimal kinodynamic motion planning problem $P$ of which the goal set is the complement set of the capture set ($\mathcal{Z}_{\text{goal}} = \mathcal{Z}_{\text{cap}}^c(\boldsymbol{z}_{\text{init}})$). 
We employ the AO-$x$ meta algorithms \cite{hauser2016asymptotically}, which are asymptotically optimal motion planners that decompose $P$ into feasible motion planning problems in the state-cost space. They iteratively find an upper bound approaching the minimum effort of escape by running as subroutines feasible kinodynamic planner $x$, such as Expansive Space Tree (EST) \cite{hsu1997path} or Rapidly-exploring Random Tree (RRT) \cite{lavalle1998rapidly} algorithms. 

The optimal motion planning problem $P = (Q, \mathcal{Z}_{\text{free}}, \mathcal{U}, \boldsymbol{z}_{\text{init}}, \mathcal{Z}_{\text{goal}}, \mathcal{Z}_b, \mathcal{U}_b, G)$ produces a trajectory $\sigma(t): [0,1] \to \mathcal{Z}_{\text{free}}$ and control $\boldsymbol{u}(t): [0,1] \to \mathcal{U}$ that minimizes the objective functional:
\begin{equation} \label{eq8}
C(\sigma) = \int_{0}^{1} Q(\sigma(t), \boldsymbol{u}(t)) dt = \int_{0}^{1} \lvert \dot{W}_{\text{ext}}(\sigma(t)) \rvert dt.
\end{equation}
Here, $Q$ denotes the incremental cost (terminal cost is $0$) following Eq. \eqref{eq2}. \redtext{$C(\sigma)$ thereby refers to the total effort of transitioning from $\sigma(0)$ to $\sigma(1)$.} $\mathcal{U}_b$ is the set of control constraints and $\boldsymbol{u} \in \mathcal{U}_b \subset \mathcal{U}$. $\mathcal{Z}_b$ refers to the set of kinematic constraints and $\boldsymbol{z} \in \mathcal{Z}_b \subset \mathcal{Z}_{\text{free}}$. The dynamics is subjected to $\boldsymbol{z}^{\prime} = G\left(\boldsymbol{z}, \boldsymbol{u}\right)$. The control input $\boldsymbol{u} = \left[\begin{array}{llll} \boldsymbol{f}^{\top} & \boldsymbol{\tau}^{\top} \end{array}\right]^{\top}$ refers to a wrench (force $\boldsymbol{f}$ applied at its CoM and torque $\boldsymbol{\tau}$) applied on the object, mimicking an external perturbation. 

Given the optimal motion planning problem $P$, a feasible motion planning problem $P_{\Bar{c}} = (\mathcal{Z}_{\text{free}} \times \mathbb{R}^{+}, \mathcal{U}, (\boldsymbol{z}_{\text{init}},0), \hat{\mathcal{Z}}_{\text{goal}}^{\Bar{c}}, \mathcal{Z}_b \times \mathbb{R}^{+}, \mathcal{U}_b, \hat{G})$ is solved at each iteration. $P_{\Bar{c}}$ augments the state in $P$ by an auxiliary cost variable, which measures the accumulated cost from the root $\boldsymbol{z}_{\text{init}}$, i.e. cost-to-come. Here, $\hat{\mathcal{Z}}_{\text{goal}}^{\Bar{c}} = \{(\boldsymbol{z}, c) : \boldsymbol{z} \in \mathcal{Z}_{\text{goal}}, c \in [0,\Bar{c}]\}$ denotes the augmented goal set with the range of cost space upper-bounded by $\Bar{c}$, and $\hat{G}$ refers to the augmented dynamics
\begin{equation} \label{eq9}
\hat{\boldsymbol{z}}^{\prime}
= \left[\begin{array}{l}
\boldsymbol{z}^{\prime} \\
{c}^{\prime}
\end{array}\right] 
= \left[\begin{array}{l}
G(\boldsymbol{z}, \boldsymbol{u}) \\
Q(\boldsymbol{z}, \boldsymbol{u})
\end{array}\right].
\end{equation}
We solve $P$ by converting it to a series of feasible motion planning subproblems $P_{\Bar{c}}$ and iteratively lowering the cost space upper bound $\Bar{c}$ utilizing a well-behaved, probabilistically complete feasible motion planner $A$ (EST or RRT). 
Informally, a planning algorithm is well-behaved \cite{hauser2016asymptotically}
if it can determine a feasible solution in finite time if it exists, and incrementally improves the cost of any found solution by a nonneglegible fraction in each further iteration.

Specifically, we first run the subroutine algorithm $A(P_\infty)$ with infinite cost space upper bound $\Bar{c} = \infty$ to find the first escape path $\sigma_0$ by growing a tree rooted at $\boldsymbol{z}_{\text{init}}$ (Algo. \ref{algo-escape}, Line 2). The tree expands a new node $\boldsymbol{z}^{\prime}$ by enforcing a randomized control $\boldsymbol{u}$ based on algorithm-specific strategies from an existing node $\boldsymbol{z}$, complying with the system dynamics $G$. The forward integration is performed in a physical simulator that allows rebounding after a collision. A new extension is valid if $\boldsymbol{z}^{\prime}$ abides by the augmented kinematic constraints $(\boldsymbol{z}^{\prime}, c(\boldsymbol{z}^{\prime})) \in \mathcal{Z}_b \times \mathbb{R}^{+}$.
If a path does not exist, it indicates the effort of escape is infinite $\Omega_{\text{esc}}\left(\boldsymbol{z}_{\text{init}}\right) = \infty$ (Line 3-5). Once we found a path $\sigma_0$ with cost $c_0 = C(\sigma_0)$, we solve a second subproblem $P_{c_0}$ (Line 8) that shrinks the cost space by lowering the upper bound from $\infty$ to $c_0$. The tree in the previous subproblem is cached and optionally pruned (discarding the nodes with costs higher than $c_0$) such that we do not grow a new tree from scratch. $A(P_{c_0})$ results in a new escape path $\sigma_1$ with cost $c_1$ (Line 9), and so on and so forth. Given the well-behavedness of $A$, the cost $c_i$ approaches the optimal cost $c^*$ as the total number of iterations approaches infinity. Therefore, we can approximately upper-bound the minimum effort of escape $\Omega_{\text{esc}}\left(\boldsymbol{z}_{\text{init}}\right)$ in finitely many ($n$) iterations. In practice, we consider EST or RRT as the subroutine algorithms $A$. RRT randomly samples a state in the state space and extends toward the sampled state from the nearest neighbor node in the tree. EST prioritizes extensions in the area with less density of existing nodes in the tree.

\begin{algorithm}[tp]
\small
\begin{spacing}{1.1}
\text{Initialize}, $P_{\infty}$, $\mathcal{Z}_{\text{cap}}(\boldsymbol{z}_{\text{init}})$, $\mathcal{Z}_{\text{suc}}$, $M$ \\
{\footnotesize $\triangleright$ \textcolor{blue}{\text{Obtain energy cost field by growing a tree}}} \\
$\{(\boldsymbol{z}^{0},c^0), ..., (\boldsymbol{z}^{M},c^M)\} \gets \textsc{EST}(P_{\infty}, M)$  \\

{\footnotesize $\triangleright$ \textcolor{blue}{\text{Approximate probabilistic distribution}}} \\
$c_{\text{min}} = \min_{0 \leq m \leq M}{c^{m}}$ \\
$S \gets {\sum_{m=0}^{M} e^{-\lambda(c^{m}-c_{\text{min}})}}$ \\
\For{$m=1,2, \ldots, M$}{
$L(\boldsymbol{z}^{m}) \gets \frac{1}{S} {e^{-\lambda(c^{m}-c_{\text{min}})}}$ \hfill {\footnotesize $\triangleright$ \textcolor{blue}{\textsc{eq. \eqref{eq4}.}}} \\
}
{\footnotesize $\triangleright$ \textcolor{blue}{\text{Compute scores}}} \\
$\Omega_{\text{cap}}(\boldsymbol{z}_{\text{init}}) \gets \sum_{m=0}^{M} \delta(\boldsymbol{z}^{m} \in \mathcal{Z}_{\text{cap}}(\boldsymbol{z}_{\text{init}})) \cdot L(\boldsymbol{z}^{m})$ \hfill {\footnotesize $\triangleright$ \textcolor{blue}{\textsc{eq. \eqref{eq5}.}}} \\
$\Omega_{\text{suc}}(\boldsymbol{z}_{\text{init}}) \gets \sum_{m=0}^{M} \delta(\boldsymbol{z}^{m} \in \mathcal{Z}_{\text{suc}}) \cdot L(\boldsymbol{z}^{m})$ \hfill {\footnotesize $\triangleright$ \textcolor{blue}{\textsc{eq. \eqref{eq6}.}}} \\
\textbf{Return} $\Omega_{\text{cap}}(\boldsymbol{z}_{\text{init}})$, $\Omega_{\text{suc}}(\boldsymbol{z}_{\text{init}})$
\end{spacing}
\caption{Compute Capture Scores}
\label{algo-likelihood}
\end{algorithm}

\subsection{Computing Capture Scores} \label{V2}
To approximate the capture scores, we construct an energy cost field through kinodynamic tree expansion. Achieving an evenly distributed sampling of state space around the initial state $\boldsymbol{z}_{\text{init}}$ is critical for this purpose. EST is thereby selected for its efficacy in promoting uniform distribution across the state space through inverse density weighting, unlike RRT, which may lead to uneven exploration. 
As detailed in Algo. \ref{algo-likelihood}, we consider the cost-augmented unbounded feasible motion planning problem $P_\infty$ (Line 2-3) with the same settings as in Section \ref{V1} except a sufficiently far away goal set $\mathcal{Z}_{\text{goal}} = \mathcal{Z}_{\infty} \subset \mathcal{Z}$  with unconstrained kinematics $\mathcal{Z}_b = \mathcal{Z}_{\text{free}}$. EST terminates after growing $M$ nodes in the tree. We thus obtain an energy cost field of $M$ state-cost pairs, $\{(\boldsymbol{z}^{0},c^0), ..., (\boldsymbol{z}^{M},c^M)\}$. We thereafter approximate the probabilistic distribution (Line 4-8) and compute the scores $\Omega_{\text{cap}}(\boldsymbol{z}_{\text{init}})$ and $\Omega_{\text{suc}}(\boldsymbol{z}_{\text{init}})$ (Line 9-11) following the procedure in Section \ref{subsec-likelihood}.

\section{Evaluation}
\label{sec-eval}
This section aims to validate the practicality and effectiveness of our metrics and algorithms by conducting simulation and real-world experiments. Our key assumption is that energy margins and caging analysis can effectively characterize robustness in manipulation. To verify this, we designed four simulated manipulation tasks and used our metrics to predict their robustness and task success. More specifically, we generated multiple trajectories of the tasks and recorded their system kinematic information and ground-truth labels of robustness and success. We computed the metrics offline, used them to make predictions and compared the results with the ground-truth labels. Through the experiments, we validated the reliable prediction capability of our metrics in environments with complex contact events. Furthermore, we verify the efficiency of our kinodynamic planning algorithms and the robustness against different model parameter errors. 

\ifthenelse{\boolean{includeFigures}}{
\begin{figure}
    \centering
    \includegraphics[width=1\linewidth]{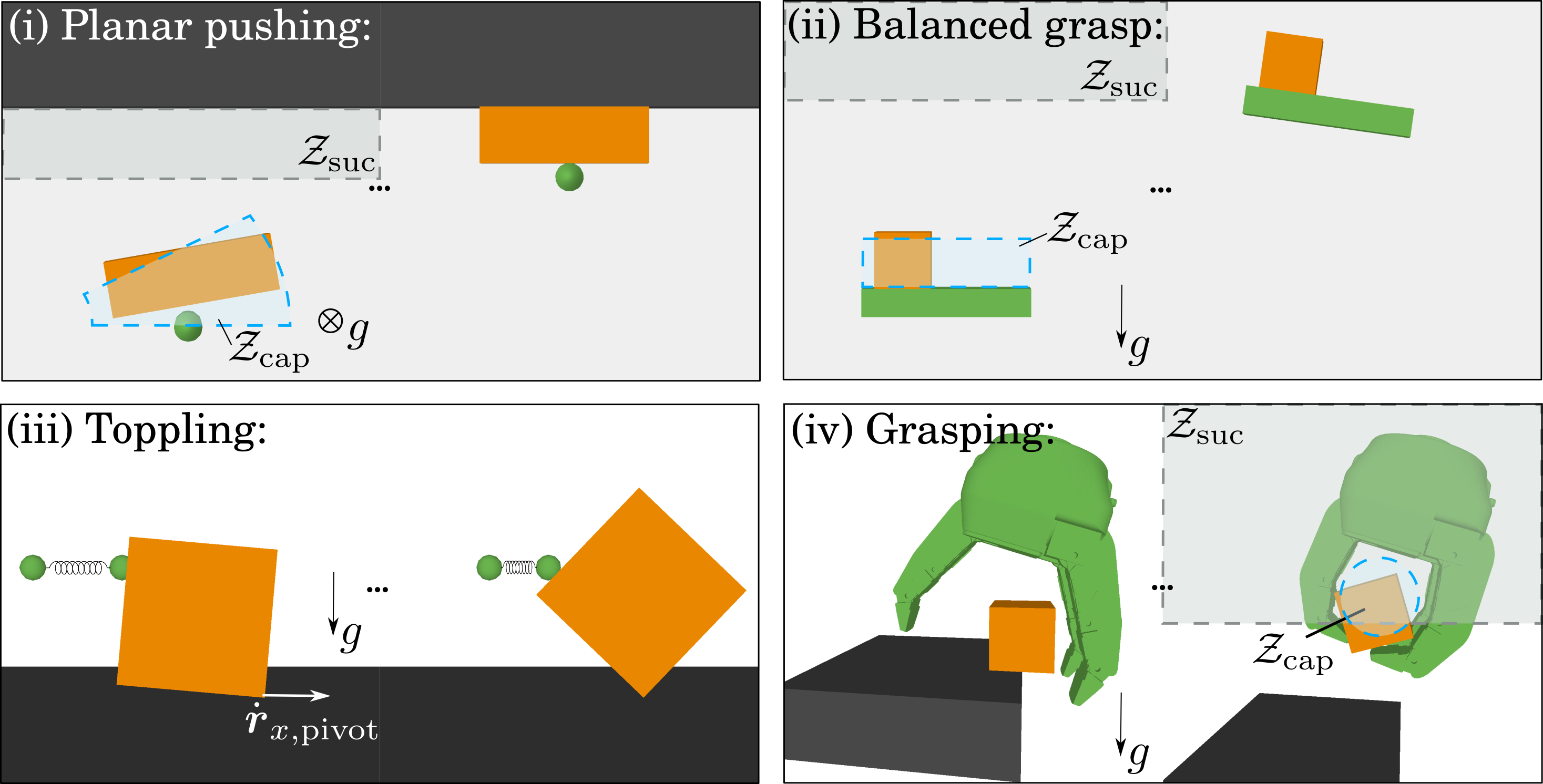}
    \caption{\purpletext{Examples of manipulation primitives. The capture sets $\mathcal{Z}_{\text{cap}}$ (blue) and task success sets $\mathcal{Z}_{\text{suc}}$ (gray) are marked.}
    }
    \label{fig-examples}
\end{figure}
}{}

\subsection{Task Description}
\label{subsec-examples}
Inspired by the manipulation primitive taxonomy in \cite{liu2014taxonomy}, \cite{ruggiero2018nonprehensile}, we design examples covering planar pushing, balanced transport, toppling, and grasping from a table. 

\textit{1) Planar Pushing}: We consider the planar manipulation problem of pushing an object on a horizontal plane towards a wall (Fig. \ref{fig-examples}-i). The task is considered successful if a state $\boldsymbol{z} \in \mathcal{Z}_{\text{suc}}$ lies in the task success set $\mathcal{Z}_{\text{suc}} = \{\boldsymbol{z} \in \mathcal{Z}: \boldsymbol{r}_x \in \mathcal{R}_{\text{suc}}\}$, and $\mathcal{R}_{\text{suc}} \subset \mathbb{R}^2$ refers to the set of object positions where the longest edge of the convex hull of the object aligns with the wall.
The capture set $\mathcal{Z}_{\text{cap}}(\boldsymbol{z}_{\text{init}}) = \{\boldsymbol{z} \in \mathcal{Z} : \boldsymbol{r}_x \in \mathcal{R}_{\text{cap}}(\boldsymbol{z}_{\text{init}})\}$ includes state $\boldsymbol{z} \in \mathcal{Z}$ such that the CoM position of the object $\boldsymbol{r}_x$ lies inside $\mathcal{R}_{\text{cap}}(\boldsymbol{z}_{\text{init}}) \subset \mathbb{R}^2$. $\mathcal{R}_{\text{cap}}(\boldsymbol{z}_{\text{init}})$ indicates the circular sector region that the end-effector will sweep over given the current instantaneous rotation center of the end-effector. 

\textit{2) Balanced Transport}: 
We aim to balance a cube on a steep slope with a rectangular support surface so that it avoids the failure mode of falling off and is transported to a target region (Fig. \ref{fig-examples}-ii). Similar to the selection of the capture set $\mathcal{Z}_{\text{cap}}(\boldsymbol{z}_{\text{init}})$ and the task success set $\mathcal{Z}_{\text{suc}}$ above, we add constraints on the object CoM position $\boldsymbol{r}_x$ such that the capture condition (maintaining on the support surface, $\mathcal{R}_{\text{cap}}(\boldsymbol{z}_{\text{init}})$) and the task success condition (reaching a goal region $\mathcal{R}_{\text{suc}}$) are satisfied.
 
\textit{3) Toppling}: 
We consider the 2D task of toppling a box on the table using a spring-like fingertip manipulator (Fig. \ref{fig-examples}-iii). The task is accomplished when the box rotates by $\pi/2$ rad pivoting on one corner in contact with the table, i.e. $\mathcal{Z}_{\text{suc}} = \{\boldsymbol{z} \in \mathcal{Z} : p(\boldsymbol{q}_x) = \pi/2 \}$, where $p(\boldsymbol{q}_x)$ is the $\mathrm{SO}(2)$ orientation of the box in Euler angle. It often fails when sliding occurs on the table contact. The capture set is defined as $\mathcal{Z}_{\text{cap}}(\boldsymbol{z}_{\text{init}}) = \{\boldsymbol{z} \in \mathcal{Z} : \|\dot{\boldsymbol{r}}_{x,\text{pivot}}\| \le 0.1 \}$, i.e. the magnitude of the sliding velocity of the table contact is smaller than 0.1 m/s.

\textit{4) Grasping from a Table}: 
The task of grasping a box lying on the desktop is considered here, using a Robotiq three-finger adaptive robot gripper with $n_r = 12$ revolute joints in the fingers (Fig. \ref{fig-examples}-iv). The task is successful when the box of edge length 10 cm is lifted above the desktop by 10 cm, while it fails if it slips and falls from the gripper. The position-controlled compliant gripper poses an external energy bound for the box to escape. Constraints on the object CoM position $\boldsymbol{r}_x$ are considered similarly as in the first two examples in selecting the sets $\mathcal{Z}_{\text{cap}}(\boldsymbol{z}_{\text{init}})$ and $\mathcal{Z}_{\text{suc}}$.

\greentext{Several other tasks with evaluations of simultaneously manipulating multiple objects, including pushing and scooping, can be found in our recent workshop contribution\footnote{\url{https://yifeidong0.github.io/assets/pdf/ICRA2024_Multi_Object.pdf}}.}
\vspace{-4pt}

\subsection{Baseline}
\label{subsec-baseline}
We compare our approach with a manipulation score in \cite{hou2019criteria}, which is designed to measure the robustness to maintain desired contact modes (sticking, sliding, etc) against disturbance forces and avoid failure contact modes (e.g. avoid slipping for sticking contacts). 
However, our energy margin approach quantifies manipulation robustness more globally and allows changes of contacts. To make a fair comparison, we design a hybrid force-based score $\Omega_{\text{force}}$ with a weighted sum of the engaging score $\Omega_{\text{engage}}$, the sticking score $\Omega_{\text{stick}}$ and a kinematic heuristic $\Omega_{\text{dist}}$ - the minimum distance between any two points on the object and the end-effector. 
$\Omega_{\text{engage}}$ uses the magnitude of the contact normal force $\lambda_{\perp}$ to evaluate how much normal disturbance forces a contact can withstand, $\Omega_{\text{engage}} = \lambda_{\perp}$. $\Omega_{\text{stick}}$ refers to the minimal amount of disturbance force required to transition the sticking contact into sliding, which is the distance between the contact force and the friction cone edges (i.e., a wrench margin away from failure):
\begin{equation}
    \Omega_{\text{stick}} = \left(\mu \lambda_{\perp} - |\lambda_{\parallelsum}| \right) \cos (\arctan \mu),
\end{equation}
where $\mu$ refers to the static friction coefficient. 

\ifthenelse{\boolean{includeFigures}}{
\begin{figure}
    \centering
    \includegraphics[width=0.99\linewidth]{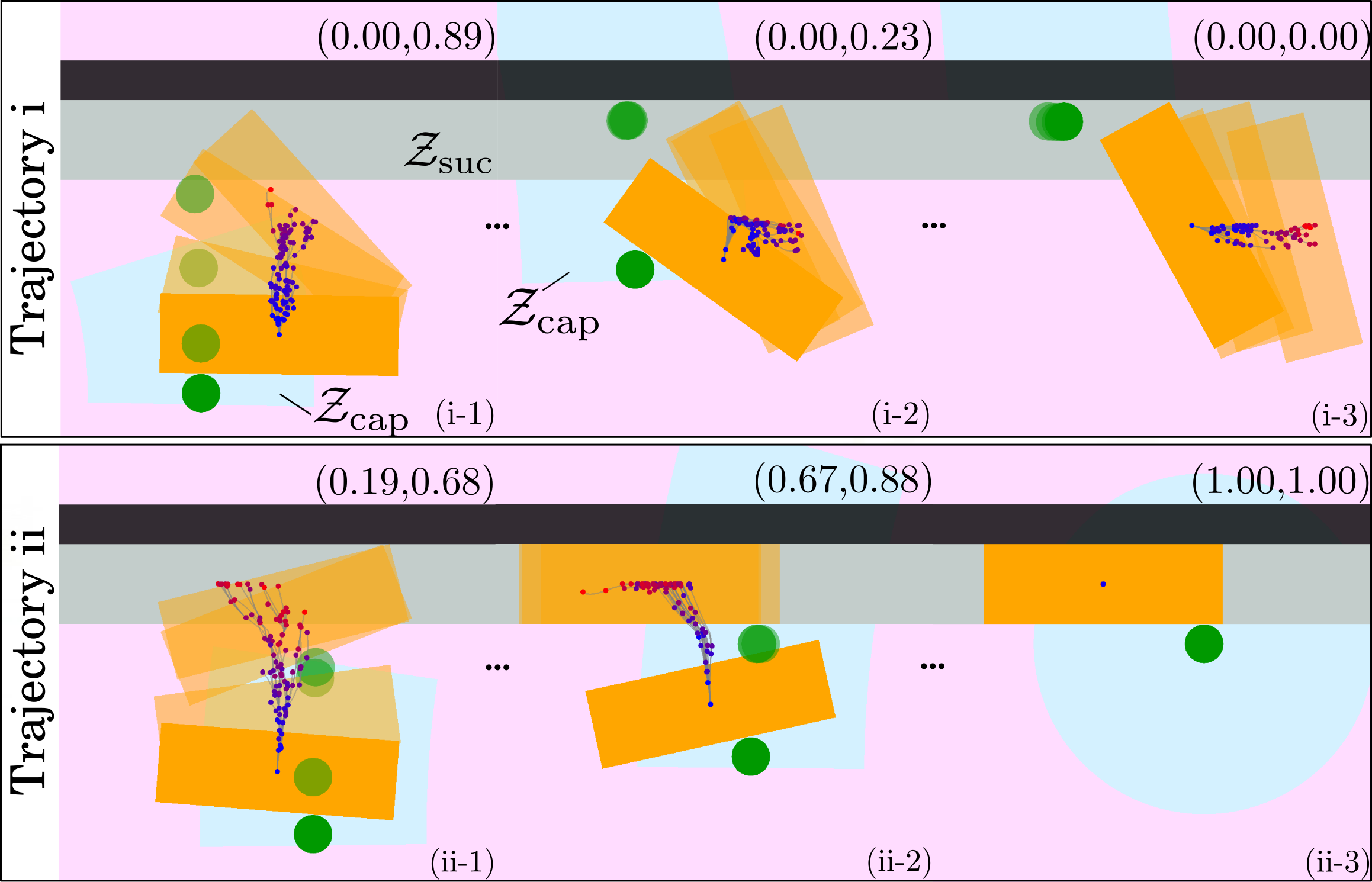}
    \caption{Qualitative evaluation for the planar pushing example. Data in the subfigures indicate $(\Omega_{\text{suc}}, \Omega_{\text{cap}})$. (i) and (ii) refer to a failed and a successful trajectory of pushing a rectangular box (yellow) to the wall (black). Three screenshots are taken along each trajectory, for which we run Algo. \ref{algo-likelihood} and visualize the expansive tree after running 100 iterations. The nodes on the tree are shown in colored dots (CoM position of the box) ranging from blue to red, implying the energy cost field. More reddish dots indicate nodes with higher cost-to-come $c$. Box and circular end-effector (green) configurations of three random nodes on the tree are visualized in partial transparency. Note the capture set $\mathcal{Z}_{\text{cap}}$ (light blue) and the task success set $\mathcal{Z}_{\text{suc}}$ (gray).}
    \label{fig-quality}
\end{figure}
}{}

\subsection{Data Generation and Implementation}
\label{subsec-practical}
We collected trajectories and per-frame system states of the 4 tasks to compute their energy margins offline and compare them with ground-truth robustness and success labels.

\textit{1) Software}: 
\greentext{ Our code can be found here\footnote{\url{https://github.com/yifeidong0/EnergyMarginDynCage}}. 
Pybullet served as the platform for forward simulation. All experiments are performed on an Intel Core i9-12900H processor with 14 cores and speeds up to 5.0 GHz.}

\textit{2) Ground-truth Data Generation}: 
In the simulation, we generated and recorded 50 trajectories for each of the four examples in Fig. \ref{fig-examples} by randomizing the initial states of the object and the end-effector, the friction coefficients, etc. 10 frames ($K=10$) are evenly selected from each trajectory and the system states across the trajectory $\boldsymbol{z}_k, k \in \{1,...,K\}$ are recorded. As an instance, two of such scripted trajectories with three states each are illustrated in Fig. \ref{fig-quality}. The hybrid force-based score $\Omega_{\text{force}} (\boldsymbol{z}_k)$ is computed by recording the friction coefficient $\mu$, the contact normal force $\lambda_{\perp}$ and lateral force $\lambda_{\parallelsum}$ in each frame. A trajectory is labeled $1$ (successful, e.g. Fig. \ref{fig-quality}, Trajectory 2) if the task objective is accomplished in the last recorded frame, i.e. $\boldsymbol{z}_K \in \mathcal{Z}_{\text{suc}}$, otherwise $0$ (Fig. \ref{fig-quality}, Trajectory 1); and a state in a frame $\boldsymbol{z}_k$ is labeled $1$ (captured) if the state is contained in the capture set for the current and subsequent $\hat{k}$ recorded frames $\boldsymbol{z}_{k^{\prime}} \in \mathcal{Z}_{\text{cap}}(\boldsymbol{z}_{{k}^{\prime}})$ for $k^{\prime} \in \{k,...,k+\hat{k}\}$.
We collected 108 successful trajectories and 92 failed trajectories.

\textit{3) Implementation Details}: 
To analyze the effectiveness of our proposed metrics in predicting the robustness and success of manipulation tasks, we utilized the Area Under the receiver operating characteristic Curve (AUC) and the Average Precision (AP) as evaluation tools. We run the algorithms in Section \ref{sec-algo} and obtain $\Omega_{\text{esc}}\left(\boldsymbol{z}_k\right)$, $\Omega_{\text{cap}}(\boldsymbol{z}_k)$ and $\Omega_{\text{suc}}(\boldsymbol{z}_k)$ for each recorded state $\boldsymbol{z}_{k}$ in each trajectory of the examples. The hybrid force-based score $\Omega_{\text{force}} (\boldsymbol{z}_k)$ is computed by recording the friction coefficient $\mu$, the contact normal force $\lambda_{\perp}$ and lateral force $\lambda_{\parallelsum}$ in each frame. We thereafter compare the scores with ground-truth labels using AUC and AP, where higher values indicate better performance across all possible classification thresholds. In Fig. \ref{fig-quality}, we demonstrated the expansive trees after running Algo. \ref{algo-likelihood}. The distribution of the energy cost field with respect to the capture set $\mathcal{Z}_{\text{cap}}$ and the task success set $\mathcal{Z}_{\text{suc}}$ directly indicates the corresponding quality score values and energy margins.
To demonstrate the capability of predicting task success using the capture score of success $\Omega_{\text{suc}}(\boldsymbol{z}_k)$, we designed a trajectory-level score $\Omega_{\text{suc}}(\boldsymbol{z}, {\Bar{k}})$ which is the weighted average of $\Omega_{\text{suc}}(\boldsymbol{z}_k)$ of only the last $\Bar{k}$ states with an increasing weight from $k=1$ to $k=\Bar{k}$. 
We thereby have, for each scenario, a dataset comprising 500 state-level data points $(\Omega_{\text{esc}}\left(\boldsymbol{z}_k\right), \Omega_{\text{cap}}(\boldsymbol{z}_k))$ and 50 trajectory-level data points $\Omega_{\text{suc}}(\boldsymbol{z}, {\Bar{k}})$ with corresponding labels. 


\begin{table}[t]
\centering
\resizebox{\columnwidth}{!}{
\large
\begin{tabular}{lcccc|c}
\hline
\textbf{Task} & $\Omega_{\text{cap}}$ & $\Omega_{\text{esc}}$ (\textbf{EST}) & $\Omega_{\text{esc}}$ (\textbf{RRT}) & $\Omega_{\text{force}}$ (\textbf{baseline}) & $\Omega_{\text{suc}}$ \\ 
\hline
Pushing & \textbf{0.97}/\textbf{0.99} & 0.92/0.98 & 0.92/\textbf{0.99} & 0.86/0.95 & 0.98/0.98 \\ 
Balancing & \textbf{0.98}/\textbf{0.99} & 0.93/0.98 & 0.94/0.98 & 0.94/0.96 & 0.93/0.95 \\
Toppling & \textbf{0.94}/\textbf{0.93} & 0.93/0.91 & 0.93/0.92 & 0.77/0.62 & 0.89/0.92 \\
Grasping & \textbf{1.00}/\textbf{1.00} & 0.99/0.99 & 0.97/0.99 & 0.73/0.88 & 0.99/0.99 \\ 
\hline
\end{tabular}
}
\caption{Energy margins evaluation in simulation (AUC/AP).}
\label{tab-sim}
\end{table}

\subsection{Quantitative Analysis}
\label{subsec-quantitative}
\textit{1) Overall Performance}: 
Upon examination of the dataset for each example in simulation (Fig. \ref{fig-examples}), AUC and AP were computed for the robustness predictions ($\Omega_{\text{cap}}$, $\Omega_{\text{esc}}$ and $\Omega_{\text{force}}$, Table \ref{tab-sim} left) and the success predictions ($\Omega_{\text{suc}}$, Table \ref{tab-sim} rightmost column). The results indicate a high level of predictive capability of our metrics. The AUC/AP values for the robustness predictions using our methods ($\Omega_{\text{cap}}$, $\Omega_{\text{esc}}$) consistently exceed 0.9 across all examples, demonstrating strong discriminative power over the baseline $\Omega_{\text{force}}$. Here, $\Omega_{\text{esc}}$ is computed by employing both RRT and EST as subroutine algorithms $A$ for Algo. \ref{algo-escape}. The values for the success predictions given $\Bar{k} = 5$ indicate reliable predictions of successful trajectories using the score $\Omega_{\text{suc}}$ we proposed.

\ifthenelse{\boolean{includeFigures}}{
\begin{figure}
    \centering
\begin{tikzpicture}

\definecolor{darkgray176}{RGB}{176,176,176}
\definecolor{darkorange25512714}{RGB}{255,127,14}
\definecolor{green}{RGB}{0,128,0}
\definecolor{steelblue31119180}{RGB}{31,119,180}

\begin{axis}[
width=0.8\linewidth, 
height=0.53\linewidth, 
log basis x={10},
tick align=inside,
tick pos=left,
xlabel={\# Iterations},
xmin=7.94328234724282, xmax=1258.92541179417,
xmode=log,
xtick style={color=black},
xtick={0.1,1,10,100,1000,10000,100000},
xticklabels={
  \(\displaystyle {10^{-1}}\),
  \(\displaystyle {10^{0}}\),
  \(\displaystyle {10^{1}}\),
  \(\displaystyle {10^{2}}\),
  \(\displaystyle {10^{3}}\),
  \(\displaystyle {10^{4}}\),
  \(\displaystyle {10^{5}}\)
},
ylabel={AUC / AP of success ratio},
legend pos=south east,
ymin=0.786945093931688, ymax=0.952401064659289,
ytick style={color=black},
xlabel style={at={(axis description cs:0.5,0.07)}, anchor=north,font=\footnotesize},
ylabel style={at={(axis description cs:0.09,.5)},anchor=south,font=\footnotesize},
legend style={font=\footnotesize},
ticklabel style={font=\footnotesize}, 
]
\path [draw=steelblue31119180, semithick]
(axis cs:10,0.802516673338672)
--(axis cs:10,0.844149993327995);

\path [draw=steelblue31119180, semithick]
(axis cs:30,0.826411010564593)
--(axis cs:30,0.913588989435407);

\path [draw=steelblue31119180, semithick]
(axis cs:100,0.880239322565748)
--(axis cs:100,0.926427344100918);

\path [draw=steelblue31119180, semithick]
(axis cs:1000,0.891391414350147)
--(axis cs:1000,0.921941918983186);

\addplot [semithick, steelblue31119180, mark=*, mark size=3, mark options={solid}]
table {%
10 0.823333333333333
30 0.87
100 0.903333333333333
1000 0.906666666666667
};
\addlegendentry{\footnotesize{AUC}}

\addplot [semithick, darkorange25512714, mark=square*, mark size=3, mark options={solid}]
table {%
10 0.823333333333333
30 0.896666666666667
100 0.926666666666667
1000 0.933333333333333
};
\addlegendentry{AP}

\addplot [semithick, green, mark=triangle*, mark size=3, mark options={solid}]
table {
10 7
30 7
};
\addlegendentry{time}

\addplot [semithick, steelblue31119180, mark=-, mark size=5, mark options={solid}, only marks]
table {%
10 0.802516673338672
30 0.826411010564593
100 0.880239322565748
1000 0.891391414350147
};
\addplot [semithick, steelblue31119180, mark=-, mark size=5, mark options={solid}, only marks]
table {%
10 0.844149993327995
30 0.913588989435407
100 0.926427344100918
1000 0.921941918983186
};
\path [draw=darkorange25512714, semithick]
(axis cs:10,0.794465819873852)
--(axis cs:10,0.852200846792814);

\path [draw=darkorange25512714, semithick]
(axis cs:30,0.873572655899082)
--(axis cs:30,0.919760677434252);

\path [draw=darkorange25512714, semithick]
(axis cs:100,0.911391414350147)
--(axis cs:100,0.941941918983186);

\path [draw=darkorange25512714, semithick]
(axis cs:1000,0.921786327949541)
--(axis cs:1000,0.944880338717126);

\addplot [semithick, darkorange25512714, mark=-, mark size=5, mark options={solid}, only marks]
table {%
10 0.794465819873852
30 0.873572655899082
100 0.911391414350147
1000 0.921786327949541
};
\addplot [semithick, darkorange25512714, mark=-, mark size=5, mark options={solid}, only marks]
table {%
10 0.852200846792814
30 0.919760677434252
100 0.941941918983186
1000 0.944880338717126
};

\end{axis}

\begin{axis}[
width=0.8\linewidth, 
height=0.53\linewidth, 
tick align=inside,
axis y line=right,
log basis x={10},
xmin=7.94328234724282, xmax=1258.92541179417,
xmode=log,
xtick pos=left,
xtick style={color=black},
xtick={0.1,1,10,100,1000,10000,100000},
xticklabels={
  \(\displaystyle {10^{-1}}\),
  \(\displaystyle {10^{0}}\),
  \(\displaystyle {10^{1}}\),
  \(\displaystyle {10^{2}}\),
  \(\displaystyle {10^{3}}\),
  \(\displaystyle {10^{4}}\),
  \(\displaystyle {10^{5}}\)
},
ylabel={Run-time per data (s)},
ymin=-0.0638573851488755, ymax=1.85520595047108,
legend pos=south east,
ytick pos=right,
ytick style={color=black},
yticklabel style={anchor=west},
axis line style=-, 
xlabel style={at={(axis description cs:0.5,-0.05)}, anchor=north,font=\footnotesize},
ticklabel style={font=\footnotesize}, 
ylabel style={at={(axis description cs:1.00,.5)},anchor=south,font=\footnotesize},
legend style={font=\footnotesize},
]
\path [draw=green, semithick]
(axis cs:10,0.0233727664702135)
--(axis cs:10,0.0240005668631198);

\path [draw=green, semithick]
(axis cs:30,0.0604664912685498)
--(axis cs:30,0.0612268420647835);

\path [draw=green, semithick]
(axis cs:100,0.183222665742258)
--(axis cs:100,0.204377334257742);

\path [draw=green, semithick]
(axis cs:1000,1.73170420114801)
--(axis cs:1000,1.76797579885199);

\addplot [semithick, green, mark=-, mark size=5, mark options={solid}, only marks]
table {%
10 0.0233727664702135
30 0.0604664912685498
100 0.183222665742258
1000 1.73170420114801
};
\addplot [semithick, green, mark=-, mark size=5, mark options={solid}, only marks]
table {%
10 0.0240005668631198
30 0.0612268420647835
100 0.204377334257742
1000 1.76797579885199
};
\addplot [semithick, green, mark=triangle*, mark size=3, mark options={solid}]
table {%
10 0.0236866666666667
30 0.0608466666666667
100 0.1938
1000 1.74984
};
\end{axis}

\end{tikzpicture}
    \caption{Time efficiency and performance (AUC/AP values for $\Omega_{\text{suc}}(\boldsymbol{z}, {\Bar{k}})$) of Algo. \ref{algo-likelihood} in a planar pushing task.
    }
    \label{fig-time}
\end{figure}
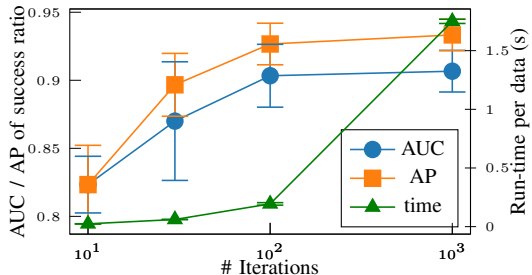
}{}

\textit{2) Algorithm Efficiency}: 
We examined the time efficiency of Algo. \ref{algo-likelihood} with respect to the number of iterations and the task success predictions (AUC/AP values for $\Omega_{\text{suc}}(\boldsymbol{z}, {\Bar{k}})$) as an ablation study. Fig. \ref{fig-time} illustrates that it takes about 1.75 sec to finish 1000 iterations in the simulated planar pushing task (Fig. \ref{fig-examples}-i). The AUC/AP values increase given more iterations of the algorithm, which is expected because of more nodes on the tree and better state space coverage. AUC/AP values reach 0.9 given only 100 iterations and about 0.2 sec of runtime, which confirms the time efficiency of our algorithm.

\textit{3) Robustness against Different Model Parameter Errors}: 
A study was conducted (in Fig. \ref{fig-perturb}) on the robustness of our algorithms given different model parameter errors, such as the positions ${\boldsymbol{r}_x}, {\boldsymbol{r}_y}$ and the velocities $\dot{\boldsymbol{x}}, \dot{\boldsymbol{y}}$ of the object and the end-effector, the contact forces $\lambda_{\parallelsum}$, $\lambda_{\perp}$, the friction coefficient $\mu$, etc. For instance, we simulated perturbed friction coefficient $\hat{\mu}$ by adding a perturbation ${\mu}_e$ randomly, uniformly sampled from the interval $[0,{e}_{\text{max}}]$ on the true values $\mu$, where ${e}_{\text{max}}$ is the maximal perturbation, i.e. $\hat{\mu} = {\mu}_e + \mu$. 

For the simulated planar pushing task, we randomized the perturbed values $\hat{\mu}$ given increasing thresholds ${e}_{\text{max}} \in \{0, {e}^1_{\text{max}}, ..., {e}^s_{\text{max}}\}$ for each trajectory $\boldsymbol{z}$. We then ran Algo. \ref{algo-likelihood} and obtained the capture score $\Omega_{\text{cap}}(\boldsymbol{z}_k)$ and the AP values with respect to scaled maximal thresholds $\Bar{e}_{\text{max}} \in \{0, {e}^1_{\text{max}}/{e}^s_{\text{max}}, ..., 1\}$ given the ground-truth labels (Fig. \ref{fig-perturb}). We generated the perturbed positions $\hat{\boldsymbol{r}}_x, \hat{\boldsymbol{r}}_y$ and the perturbed velocities $\hat{\dot{\boldsymbol{x}}}, \hat{\dot{\boldsymbol{y}}}$ for each state-level data point ${\boldsymbol{z}}_k$ and the corresponding AP curves of velocity and position. 
Similarly, the perturbed contact forces $\hat{\lambda}_{\parallelsum}$, $\hat{\lambda}_{\perp}$ are generated for the force-based score $\Omega_{\text{force}}(\boldsymbol{z}_k)$ with the AP curve of force. 
Since different types of error have different units, we scaled their maximal thresholds ${e}^s_{\text{max}}$ to $\Bar{e}_{\text{max}}$ so they can be shown on one figure.
While higher modeling errors do affect the prediction capability, our methods (the friction, velocity, and position curves) demonstrate better robustness against model parameter errors than the baseline (the force curve). This is because our method inherently captures many hard-to-model effects through rollout and energy analysis. 

\ifthenelse{\boolean{includeFigures}}{
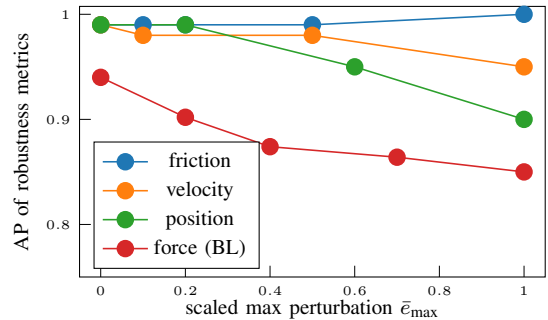
\begin{figure}
    \centering
\begin{tikzpicture}

\definecolor{crimson2143940}{RGB}{214,39,40}
\definecolor{darkgray176}{RGB}{176,176,176}
\definecolor{darkorange25512714}{RGB}{255,127,14}
\definecolor{forestgreen4416044}{RGB}{44,160,44}
\definecolor{steelblue31119180}{RGB}{31,119,180}

\begin{axis}[
width=0.9\linewidth, 
height=0.6\linewidth, 
tick pos=left,
xlabel={scaled max perturbation $\Bar{e}_{\text{max}}$},
tick align=inside,
legend pos=south west,
xmin=-0.05, xmax=1.05,
ticklabel style={font=\footnotesize}, 
xtick style={color=black,font=\footnotesize},
ylabel={AP of robustness metrics},
ymin=0.75, ymax=1.0075,
ytick style={color=black,font=\footnotesize},
xlabel style={at={(axis description cs:0.5,0.07)}, anchor=north,font=\footnotesize},
ylabel style={at={(axis description cs:0.09,.5)},anchor=south,font=\footnotesize},
legend style={font=\footnotesize},
]
\addplot [semithick, steelblue31119180, mark=*, mark size=3, mark options={solid}]
table {%
0 0.99
0.1 0.99
0.2 0.99
0.5 0.99
1 1
};
\addlegendentry{friction}

\addplot [semithick, darkorange25512714, mark=*, mark size=3, mark options={solid}]
table {%
0 0.99
0.1 0.98
0.5 0.98
1 0.95
};
\addlegendentry{velocity}

\addplot [semithick, forestgreen4416044, mark=*, mark size=3, mark options={solid}]
table {%
0 0.99
0.2 0.99
0.6 0.95
1 0.9
};
\addlegendentry{position}

\addplot [semithick, crimson2143940, mark=*, mark size=3, mark options={solid}]
table {%
0 0.94
0.2 0.902
0.4 0.874
0.7 0.864
1 0.85
};
\addlegendentry{force (BL)}

\end{axis}

\end{tikzpicture}
    \caption{Robustness evaluation of Algo. \ref{algo-likelihood} under various modeling errors: This plot shows the algorithm performance against different estimation errors of various types, including friction coefficient ($\mu$), velocity ($\dot{\boldsymbol{x}}, \dot{\boldsymbol{y}}$), relative position (${\boldsymbol{r}_x}, {\boldsymbol{r}_y}$), and contact forces ($\lambda_{\parallelsum}, \lambda_{\perp}$). The graph demonstrates how AP values for the capture score $\Omega_{\text{cap}}$ (blue, orange and green) and the force-based baseline (BL) score $\Omega_{\text{force}}$ (red) vary given increasing maximal error thresholds ($\Bar{e}_{\text{max}}$).
    }
    \label{fig-perturb}
\end{figure}
}{}

\subsection{Real-World Experiment}
\label{subsec-realworld}
\textit{1) Setup and Data Collection}: 
We conducted real-world experiments using an Interbotix WidowX-200 robot arm (Fig. \ref{fig-physical}-i) for the planar pushing task of various geometric shapes of 3D-printed objects (five shapes) and end-effectors (two shapes) (Fig. \ref{fig-physical}-ii). The arm has a position error between 5-8 mm, providing a testbed to evaluate our scores under system uncertainty as in Section \ref{subsec-quantitative}-3. 
We used a joystick to manually control the end-effector positions to perform the planar pushing task in the horizontal desktop plane. Visual markers were attached to the object and the end-effector to record their poses over time using a Realsense D415 camera. 

We considered ten combinations of objects and end-effectors. For each combination, we collected 12 trajectories, with 100 recorded states for each trajectory. The velocities were computed from the poses and the frequency of the recording (33 Hz). A human observer provided success labels of the task objective. Failed (Fig. \ref{fig-physical}-iv) and successful (Fig. \ref{fig-physical}-v) trajectories are illustrated. Given the recorded states and system kinematic and dynamic parameters (mass, moment of inertia, friction coefficients, geometric information, etc.), the trajectories were replicated in simulation (Fig. \ref{fig-physical}-iii). We use the simulation to compute energy margins offline. 

\ifthenelse{\boolean{includeFigures}}{
\begin{figure}
    \centering
    \includegraphics[width=1\linewidth]{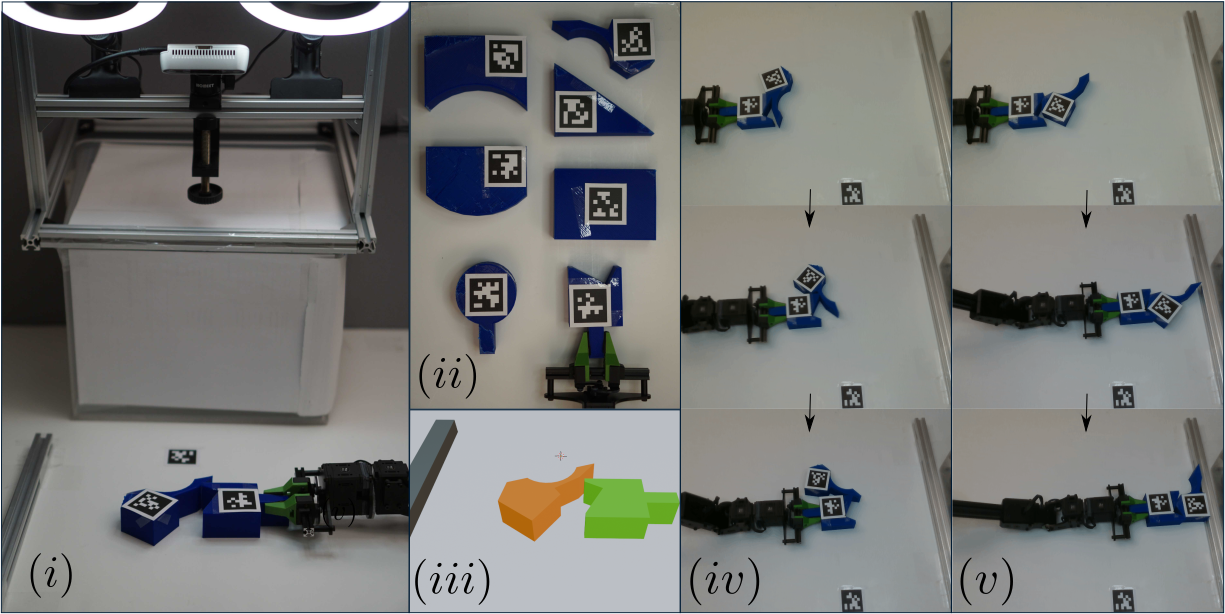}
    \caption{Real-world experiments settings. (i) The planar pushing task using an Interbotix WidowX-200 robot arm with a top-view camera. (ii) 5 3D-printed objects and 2 end-effectors. (iii) Replication in the simulation of the scene in (i) for computing the energy margins. (iv,v) Screenshots of a failed/successful trajectory from the camera view.
    }
    \label{fig-physical}
\end{figure}
}{}

\textit{2) Geometry Effect Evaluation}: 
Statistical analysis results of the AUC/AP values for the scores $\Omega_{\text{cap}}(\boldsymbol{z}_k)$ and $\Omega_{\text{suc}}(\boldsymbol{z}, \Bar{k})$, are summarized in Table \ref{tab-physical} following the same procedure as in the simulation (Section \ref{subsec-practical}) given $\Bar{k}=60, K=100$. The effort of escape $\Omega_{\text{esc}}$ is not discussed here due to its inferior performance compared to the capture score in Table \ref{tab-sim}. The results presented mostly exceed 0.85, which supports the discrimination and predictivity power of our scores and is in line with the results in the simulation. The performance of the capture score of success $\Omega_{\text{suc}}(\boldsymbol{z})$ of 3 of the 10 combinations of geometries (the jaw end-effector with the triangle, concave and irregular objects) is below the average though. A reason for the failure could be that the sim-to-real gap is especially sensitive to these cases with complicated geometric shapes.

\begin{table}[ht]
\vspace{5pt}
\centering
\footnotesize
\begin{tabular}{llcc}
\hline
\textbf{Object} & \textbf{End-effector} & \textbf{Robust} $\Omega_{\text{cap}}$ & \textbf{Success} $\Omega_{\text{suc}}$ \\ \hline
Rectangle & Jaw & 0.85 / 0.89 & 0.94 / 0.95 \\
 & Round & 0.89 / 0.98 & 1.00 / 1.00 \\ \hline
Triangle & Jaw & 0.94 / 0.99 & 0.69 / 0.62 \\
 & Round & 0.93 / 0.99 & 0.94 / 0.93 \\ \hline
Convex & Jaw & 0.91 / 0.92 & 1.00 / 1.00 \\
 & Round & 0.84 / 0.81 & 0.91 / 0.88 \\ \hline
Concave & Jaw & 0.93 / 0.96 & 0.67 / 0.78 \\
 & Round & 0.97 / 0.99 & 0.94 / 0.93 \\ \hline
Irregular & Jaw & 0.95 / 0.99 & 0.47 / 0.36 \\
 & Round & 0.94 / 0.99 & 0.93 / 0.87 \\ \hline
\end{tabular}
\caption{Real-world experiments result (AUC/AP).}
\label{tab-physical}
\end{table}
\section{Conclusion}
In this paper, we propose to quantify manipulation robustness through energy margins to failure and compute them using kinodynamic motion planning algorithms. We believe this approach is a step towards robustness characterization of general dexterous manipulation. Our approach currently has certain limitations, such as reliance on dynamic rollouts in simulation, which requires system modeling and causes computational costs. Additionally, our analysis is confined to rigid objects and utilizes an empirically defined capture set. Looking ahead, we aim to improve our method by overcoming these limitations and applying our robustness characterization in motion planning for general manipulation.

\section*{Acknowledgment}
We extend our gratitude to Robert Gieselmann and Rafael Cabral for their discussions on this work. 


\ifCLASSOPTIONcaptionsoff
  \newpage
\fi



%


\bibliographystyle{bibliography/IEEEtran}
\bibliography{bibliography/references}
%
\end{document}